\pdfoutput=1
\documentclass{fairmeta}
\usepackage{amsmath,amssymb}
\usepackage{lmodern}
\usepackage{xspace,colortbl,tabularx}
\setcitestyle{numbers,square,comma}
\newsavebox{\arxivtablebox}
\newcommand{\arxivtablestyle}{%
  \normalfont\normalsize
  \renewcommand{\arraystretch}{1.14}%
  \setlength{\tabcolsep}{5pt}%
  \setlength{\heavyrulewidth}{0.7pt}%
  \setlength{\lightrulewidth}{0.4pt}%
  \setlength{\cmidrulewidth}{0.3pt}%
  \arrayrulecolor{black!70}%
  \captionsetup{skip=4pt}%
}
\newcommand{\method}{\textnormal{GenRA}\xspace}

\title{GenRA: Ultra-Low-Bitrate \textcolor{metablue}{Gen}erative Video Compression via \textcolor{metablue}{R}epresentation \textcolor{metablue}{A}daptation}
\author[\ddagger]{Jiancheng Zhao}
\author[\ddagger]{Jixiang Luo}
\author{Haibin Huang}
\author{Chi Zhang}
\author[*]{Xuelong Li}
\affiliation{Institute of Artificial Intelligence (TeleAI), China Telecom}
\metadata[Keywords]{generative video compression, AI Flow}
\metadata[\textsuperscript{*}Correspondence to]{Xuelong Li (\href{mailto:xuelong_li@ieee.org}{\texttt{xuelong\_li@ieee.org}})\par
\textsuperscript{\ensuremath{\ddagger}}Equal contribution. This work was conducted at TeleAI.}
\abstract{Preserving fine visual details in ultra-low-bitrate video compression remains challenging due to the loss of critical information. Recent generative video compression methods exploit generative priors to improve perceptual quality.
These pretrained generative representations are generally not optimized for compression. Some approaches therefore use an additional learned codec to map them into a separate latent space for compression. To this end, we propose \method, which adapts pretrained video representations for direct quantization and entropy coding without an additional learned codec. Specifically, we use vector quantization to adapt a pretrained video VAE into a discrete video tokenizer. A lightweight categorical autoregressive entropy model combines local and global context to model the probabilities of the resulting token indices.
We jointly optimize the tokenizer and entropy model using a differentiable rate surrogate, allowing the encoder to adjust token assignments based on coding cost and reconstruction quality. Experiments on UVG, MCL-JCV, and HEVC Class B show average bitrate savings of 84.1\% and 82.3\% over GLC-Video in terms of LPIPS and DISTS, respectively.
}
\hypersetup{pdftitle={GenRA: Ultra-Low-Bitrate Generative Video Compression via Representation Adaptation},pdfauthor={Jiancheng Zhao, Jixiang Luo, Haibin Huang, Chi Zhang, Xuelong Li},pdfkeywords={generative video compression, AI Flow}}
\begin{document}
\maketitle
\begin{figure}[tbp]
\centering
\includegraphics[width=0.78\linewidth]{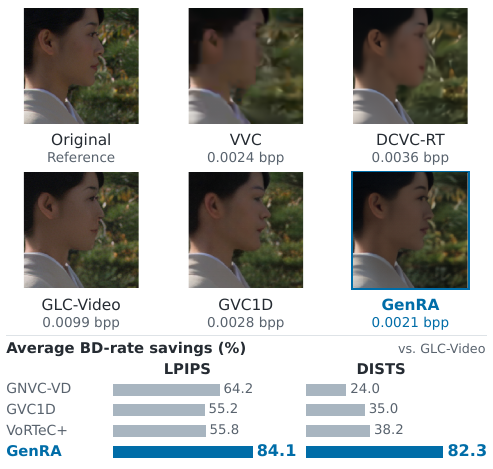}
\caption{Top: visual comparisons. Bottom: average BD-rate savings over GLC-Video~\cite{qi2025glc} across UVG, MCL-JCV, and HEVC Class B; \method saves 84.1\% (LPIPS) and 82.3\% (DISTS).}
\label{fig:teaser}
\end{figure}

\section{Introduction}
\label{sec:introduction}

Video compression at ultra-low bitrates remains challenging, as limited bit budgets often result in blurred textures and the loss of fine details.
Generative Video Compression (GVC) addresses this challenge by exploiting pretrained generative models to reconstruct perceptually plausible details~\cite{chen2026gvc}.
These large-scale pretrained generative models~\cite{minimax2026h3,wan2025,ltxvideo2025} provide rich visual priors that have shown promise in lots of areas such as video restoration~\cite{vividvr2025,pretrainedrestoration2026,fastvr2026}, editing~\cite{revideo2024,videopainter2025,vace2025}, and super-resolution~\cite{star2025,dove2025,flashvsr2025}.
Leveraging these generative models, recent GVC methods~\cite{xue2026generative,gao2026zerogvc,wang2026group,ling2026free,zeng2026gvcc} achieve superior perceptual quality at ultra-low bitrates, outperforming traditional~\cite{sullivan2012overview,bross2021overview} and non-generative learned video compression methods~\cite{jia2025towards,li2024neural}.

Pretrained generative representations are generally not optimized for compression~\cite{yao2025vavae,ye2026infotok}.
Many GVC methods therefore use an additional learned codec~\cite{jia2025towards,li2026ultra} to map these representations into a separate latent space for compression, followed by video reconstruction using their perceptual decoders~\cite{mao2026generative,xue2026generative}.
Although these methods improve perceptual quality, the additional codec may also introduce another source of reconstruction loss.
Recently, some approaches have begun to explore coding directly in generative latent spaces~\cite{zhou2025tvc,li2026progvc}.
However, adapting the pretrained video representations themselves to balance bitrate and reconstruction quality has yet to be systematically explored.

We propose \method, a novel framework for generative video compression via representation adaptation, which enables direct quantization and entropy coding of pretrained video representations~\cite{minimax2026h3} without an additional learned codec.
To this end, \method is built around three key design choices.
We first use vector quantization~\cite{oord2017vqvae} to convert the continuous latents into token indices for entropy modeling.
The entropy model then estimates the resulting token probabilities using both local and global context.
To reduce sequential decoding steps, we divide the tokens into groups and predict the tokens within each group in parallel.
When trained separately, the entropy model cannot guide the encoder toward token assignments with lower coding costs.
Following RDVQ~\cite{jiang2026rdvq}, we use a differentiable rate surrogate that trains the entropy model on hard token assignments and propagates rate gradients to the encoder through soft assignments.
As shown in Fig.~\ref{fig:teaser}, \method achieves average bitrate savings of 84.1\% and 82.3\% over GLC-Video~\cite{qi2025glc} in terms of LPIPS and DISTS, respectively, across three representative benchmarks UVG~\cite{mercat2020uvg}, MCL-JCV~\cite{wang2016mcl}, and HEVC-B.

Our contributions are summarized as follows:
\begin{itemize}
    \item We propose \method, which adapts pretrained video representations through vector quantization and joint rate--distortion optimization for direct compression in the generative latent space.
    \item We develop an entropy model that combines local and global context to estimate token probabilities and uses window-parallel grouped prediction to reduce sequential decoding steps.
    \item We demonstrate state-of-the-art perceptual rate--distortion performance on UVG~\cite{mercat2020uvg}, MCL-JCV~\cite{wang2016mcl}, and HEVC Class B.
\end{itemize}

\section{Related Work}
\label{sec:related-work}

\paragraph{Generative video compression.}
GVC exploits learned generative priors to improve perceptual quality at low bitrates~\citep{chen2026gvc,an2026aiflow,chen2026generative}.
Early approaches use perceptual and adversarial training to reconstruct realistic details~\citep{yang2022perceptual,mentzer2022neural}, while more recent methods incorporate pretrained video generative models to improve perceptual quality and temporal coherence~\citep{mao2026generative,ding2026cgvc,gao2026zerogvc,ling2026free}.
For example, GNVC-VD~\citep{mao2026generative} uses a contextual latent codec based on DCVC-RT~\citep{jia2025towards} to compress representations from the pretrained Wan2.1~\citep{wan2025} VAE, then refines the decoded latents with a video diffusion model.
In contrast, \method adapts the VAE representations of MiniMax-H3~\citep{minimax2026h3} for direct quantization and entropy coding without an additional learned codec.

\paragraph{Generative representation coding.}
Compression in generative latent spaces has been widely explored.
Beyond pretrained video generative models, GLC-Video~\citep{qi2025glc} exploits the generative capabilities of a VQ-VAE~\citep{oord2017vqvae}, while GVC1D~\citep{zheng2026gvc1d} uses a pretrained TA-TiTok tokenizer~\citep{kim2025democratizing} to represent each frame with a compact set of one-dimensional tokens.
However, both use frame-wise conditional coding based on previously decoded representations, which can propagate reconstruction errors across frames.
Earlier approaches exploit temporal redundancy using 3D convolutional autoencoders~\citep{habibian2019video} or Transformer-based temporal entropy models~\citep{mentzer2022vct}.
Although these methods compress video in latent space, they do not use pretrained generative priors for perceptual reconstruction at ultra-low bitrates.
Our approach combines these directions by adapting the joint spatiotemporal representations of a pretrained video VAE for quantization and entropy coding.

Recent work explores pretrained video tokenizers for compression.
TVC~\cite{zhou2025tvc} combines discrete and continuous streams from frozen Cosmos tokenizers~\cite{agarwal2025cosmos} to improve reconstruction quality.
However, its continuous stream incurs a fixed cost of approximately 0.003 bpp in their reported configuration, limiting further bitrate reduction through discrete-token masking.
ProGVC~\cite{li2026progvc} adapts the coarse-to-fine generation framework of InfinityStar~\cite{liu2026infinitystar} for progressive coding, controlling bitrate by transmitting a subset of token scales and generating the remaining scales.
\method adapts the continuous representations of the MiniMax-H3~\cite{minimax2026h3} video VAE into a single discrete spatiotemporal token stream, avoiding the auxiliary continuous-stream cost of TVC~\cite{zhou2025tvc}.
Our main rate--distortion operating points are obtained through joint tokenizer and entropy model optimization under different rate constraints, rather than by varying the number of transmitted scales as in ProGVC~\cite{li2026progvc}.

\section{\method}
\label{sec:method}

\begin{figure}[tbp]
\centering
\includegraphics[width=\linewidth]{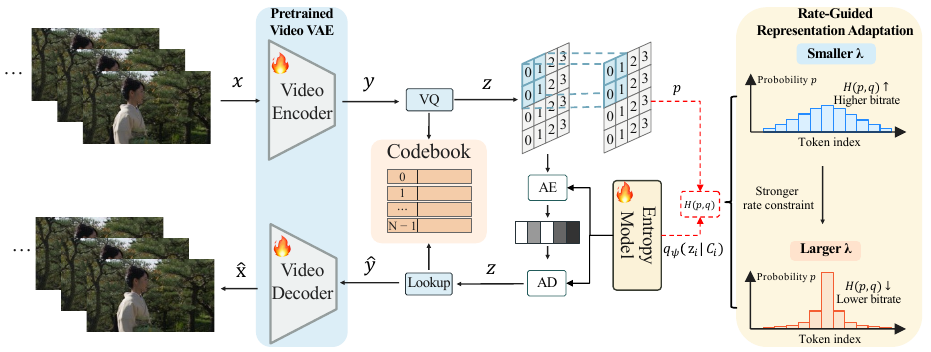}
\caption{Overview of \method.
\textsc{Left:} The encoder maps input video $\mathbf{x}$ to continuous spatiotemporal latents $\mathbf{y}$, which are vector-quantized into token indices $\mathbf{z}$.
The entropy model estimates the probabilities used to encode $\mathbf{z}$ into a bitstream.
During decoding, codebook lookup maps the recovered indices to quantized latents, and the decoder reconstructs $\hat{\mathbf{x}}$.
\textsc{Right:} A stronger rate constraint encourages the encoder to produce more concentrated token distributions that are easier to predict, resulting in lower bitrates.}
\label{fig:overview}
\end{figure}

\method adapts pretrained generative video representations for efficient compression.
As illustrated in \Cref{fig:overview}, the framework comprises a video encoder $E_{\theta}$, a vector quantizer $Q_{\mathcal{E}}$ with codebook $\mathcal{E}$, an entropy model $q_{\psi}$, and a video decoder $D_{\theta}$.

Given an input video $\mathbf{x}\in\mathbb{R}^{3\times T\times H\times W}$, the encoder produces continuous spatiotemporal latents $\mathbf{y}\in\mathbb{R}^{d\times T_y\times H_y\times W_y}$.
The vector quantizer maps these latents to token indices $\mathbf{z}\in\{0,\ldots,N-1\}^{T_y\times H_y\times W_y}$ and their corresponding codebook embeddings $\hat{\mathbf{y}}$:
\begin{equation}
\mathbf{y}=E_{\theta}(\mathbf{x}), \qquad
(\mathbf{z},\hat{\mathbf{y}})=Q_{\mathcal{E}}(\mathbf{y}).
\label{eq:encoding}
\end{equation}
Here, $N$ denotes the codebook size, $d$ is the number of latent channels, and $T_y$, $H_y$, and $W_y$ denote the temporal length, height, and width of the latent representation, respectively.
The entropy model $q_{\psi}$ estimates conditional token probabilities from previously decoded context.
These probabilities are used to encode $\mathbf{z}$ into a bitstream via arithmetic coding.
At the receiver, arithmetic decoding recovers $\mathbf{z}$, and codebook lookup retrieves the corresponding embeddings $\hat{\mathbf{y}}$.
The video decoder then reconstructs the video:
\begin{equation}
\hat{\mathbf{x}}=D_{\theta}(\hat{\mathbf{y}}).
\end{equation}

\subsection{Generative Video Representation}

Coding in a generative latent space allows us to exploit pretrained generative priors to recover perceptual details at ultra-low bitrates.
We therefore build \method on the pretrained MiniMax-H3 video VAE~\cite{minimax2026h3} and adapt its continuous latents through vector quantization (VQ)~\cite{oord2017vqvae} for subsequent entropy modeling.

Specifically, the encoder in Eq.~\ref{eq:encoding} uses causal 3D convolutions to map the input video $\mathbf{x}$ to continuous latents $\mathbf{y}$, with downsampling factors of $4$ temporally and $16$ along each spatial dimension.
The resulting continuous latents capture both spatial and temporal information.
We process the video in 17-frame chunks, concatenate the resulting latent sequences, and discard the final three latent frames, yielding
\begin{equation}
T_y=5\left\lceil\frac{T}{17}\right\rceil-3.
\end{equation}

To obtain discrete representations, we assign each continuous latent $\mathbf{y}_i$ at spatiotemporal position $i$ to the nearest codeword in a codebook $\mathcal{E}$ containing $N$ codewords of dimension $d$:
\begin{equation}
z_i=\arg\min_{v\in\{0,\ldots,N-1\}}
\lVert\mathbf{y}_i-\mathbf{e}_v\rVert_2^2.
\label{eq:vq-assignment}
\end{equation}
The resulting indices form a three-dimensional grid $\mathbf{z}$ of size $T_y\times H_y\times W_y$, which is subsequently entropy-coded into a bitstream.

\begin{figure}[tbp]
\centering
\begin{minipage}[t]{0.43\linewidth}
\centering
\vspace{0pt}
\begin{minipage}[c][3.1cm][c]{\linewidth}
\arxivtablestyle
\centering
\begin{tabular}{@{}lrr@{}}
\toprule
Tokenizer & LPIPS $\downarrow$ & DISTS $\downarrow$ \\
\midrule
H3 VAE & 0.1941 & 0.0626 \\
GenRA (VQ) & \textbf{0.1870} & \textbf{0.0474} \\
\bottomrule
\end{tabular}

\end{minipage}
\captionof{table}{Perceptual quality before and after VQ adaptation.}
\label{tab:tokenizer-adaptation}
\end{minipage}\hfill
\begin{minipage}[t]{0.51\linewidth}
\centering
\vspace{0pt}
\begin{minipage}[c][3.1cm][c]{\linewidth}
\centering
\includegraphics[height=3.1cm,width=\linewidth,keepaspectratio]{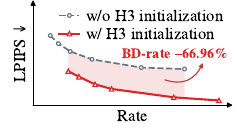}
\end{minipage}
\captionof{figure}{Effect of H3 initialization on UVG.}
\label{fig:uvg-lpips-scratch}
\end{minipage}
\end{figure}

Table~\ref{tab:tokenizer-adaptation} compares perceptual reconstruction quality before and after VQ adaptation, showing that the discrete tokenizer retains the original VAE's perceptual reconstruction capability.
Figure~\ref{fig:uvg-lpips-scratch} compares initialization from pretrained weights with training the same architecture from scratch, showing a bitrate reduction of 66.96\% with pretrained initialization.

\subsection{Autoregressive Entropy Model}
\label{sec:entropy-model}

\begin{figure}[tbp]
\centering
\includegraphics[width=0.76\linewidth]{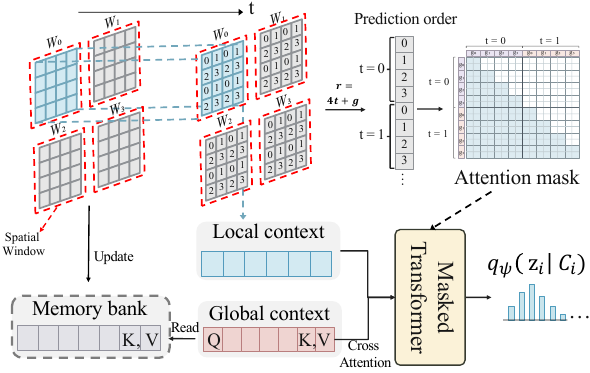}
\caption{Window-parallel autoregressive entropy modeling.
Tokens follow the frame-major group order $O(t,g)=4t+g$, with corresponding groups predicted in parallel across windows.
A masked Transformer uses local context and retrieves complementary global context from a memory bank through cross-attention.}
\label{fig:entropy-model}
\end{figure}

Directly storing $\mathbf{z}$ does not exploit its spatial and temporal dependencies.
We therefore use an entropy model to estimate token probabilities for entropy coding.
However, token-by-token prediction requires many sequential decoding steps.
As illustrated in \Cref{fig:entropy-model}, \method combines window partitioning with grouped prediction to reduce the number of sequential prediction steps.

\paragraph{Window-parallel grouped prediction.}
We process latent frames in temporal order and partition each frame into non-overlapping $8\times8$ spatial windows.
Within each window, a checkerboard pattern divides the tokens into four groups $g\in\{0,1,2,3\}$, which are predicted in the order $0 \rightarrow 1 \rightarrow 2 \rightarrow 3$.
The prediction order of group $g$ in latent frame $t$ is therefore
\begin{equation}
O(t,g)=4t+g.
\end{equation}
At each step, all tokens in the corresponding group are predicted in parallel across windows, requiring only four sequential prediction steps per latent frame.

This order determines the attention mask within each spatial window:
\begin{equation}
M_{ij}=
\begin{cases}
1, & O(t_j,g_j)<O(t_i,g_i),\\
0, & \text{otherwise},
\end{cases}
\end{equation}
where $i$ and $j$ denote query and key token positions, respectively.
Each token can therefore attend only to tokens with an earlier prediction order.

\paragraph{Local and global context.}
For each group, local context consists of previously decoded groups in the current window and decoded tokens at the same spatial window location in earlier latent frames.
A masked Transformer uses this context to model local spatiotemporal dependencies.

To complement this window-restricted context, we introduce a fixed-size global memory bank that summarizes previously decoded latent frames.
Learnable queries retrieve global context from the memory through cross-attention, providing information from other windows in previous frames to support prediction of the current group.

Together, the local and global context form the conditioning context $C_i$ for token $z_i$.
Under the grouped prediction order, the joint probability factorizes as
\begin{equation}
q_{\psi}(\mathbf{z})
=
\prod_{r=0}^{4T_y-1}
\prod_{i\in\mathcal{G}_r}
q_{\psi}(z_i\mid C_i),
\label{eq:entropy-factorization}
\end{equation}
where $\mathcal{G}_r$ denotes the token positions predicted across all windows at step $r$.

\section{Experiments}
\label{sec:experiments}

\subsection{Perceptual Rate--Distortion Performance}
\label{sec:rd-performance}

\begin{figure}[tbp]
\centering
\includegraphics[width=\linewidth]{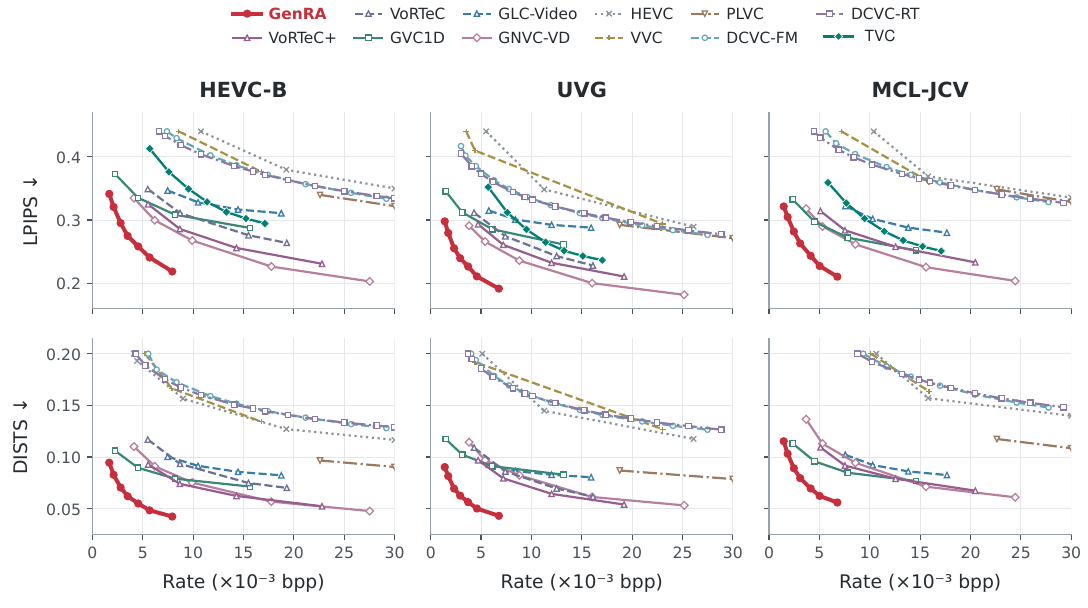}
\caption{\textbf{Perceptual rate--distortion comparison with state-of-the-art video codecs on HEVC-B, UVG, and MCL-JCV.}}
\label{fig:rd-curve}
\end{figure}
\begin{table}[tbp]
\arxivtablestyle
\centering
\sbox{\arxivtablebox}{%
\begin{tabular}{@{}lrrrrrr@{}}
\toprule
\textbf{Method} & \multicolumn{2}{c}{\textbf{HEVC-B}} & \multicolumn{2}{c}{\textbf{UVG}} & \multicolumn{2}{c}{\textbf{MCL-JCV}} \\
\cmidrule(lr){2-3}\cmidrule(lr){4-5}\cmidrule(lr){6-7}
& LPIPS & DISTS & LPIPS & DISTS & LPIPS & DISTS \\
\midrule
GLC-Video~\cite{qi2025glc} & $0.0$ & $0.0$ & $0.0$ & $0.0$ & $0.0$ & $0.0$ \\
\addlinespace[2pt]
GNVC-VD~\cite{mao2026generative} & $-61.9$ & $-43.3$ & $-72.9$ & $-13.0$ & $-57.9$ & $-15.6$ \\
GVC1D~\cite{zheng2026gvc1d} & $-51.6$ & $-61.5$ & $-53.0$ & $6.0$ & $-60.9$ & $-49.5$ \\
VoRTeC+~\cite{xia2026vortectamingfoundationflow} & $-57.9$ & $-53.6$ & $-63.7$ & $-31.9$ & $-45.7$ & $-29.0$ \\
\midrule
\textbf{GenRA (Ours)} & $\mathbf{-82.6}$ & $\mathbf{-85.0}$ & $\mathbf{-87.5}$ & $\mathbf{-83.5}$ & $\mathbf{-82.2}$ & $\mathbf{-78.3}$ \\
\bottomrule
\end{tabular}%
}
\begin{minipage}{\wd\arxivtablebox}
\caption{\textbf{Perceptual rate--distortion comparison.} BD-rate (\%) vs. GLC-Video~\cite{qi2025glc}; negative values indicate bitrate savings.}
\label{tab:main-bd-rate}
\noindent\usebox{\arxivtablebox}
\end{minipage}

\end{table}
\paragraph{Quantitative comparisons.}
\Cref{fig:rd-curve} shows perceptual rate--distortion curves on UVG~\cite{mercat2020uvg}, MCL-JCV~\cite{wang2016mcl}, and HEVC-B.
\method achieves state-of-the-art perceptual rate--distortion performance.
As summarized in \Cref{tab:main-bd-rate}, \method achieves average bitrate savings of $84.1\%$ and $82.3\%$ over GLC-Video~\cite{qi2025glc} across the three datasets, measured by BD-rate using LPIPS~\cite{zhang2018unreasonable} and DISTS~\cite{ding2020image}, respectively.
\method also outperforms GVC1D~\cite{zheng2026gvc1d}, which uses compact one-dimensional tokens, as well as GNVC-VD~\cite{mao2026generative} and VoRTeC+~\cite{xia2026vortectamingfoundationflow}, which use pretrained generative models for reconstruction.
These results support our approach of quantizing pretrained video representations and modeling their token probabilities under joint rate--distortion optimization.

\begin{figure}[tbp]
\centering
\includegraphics[width=\linewidth]{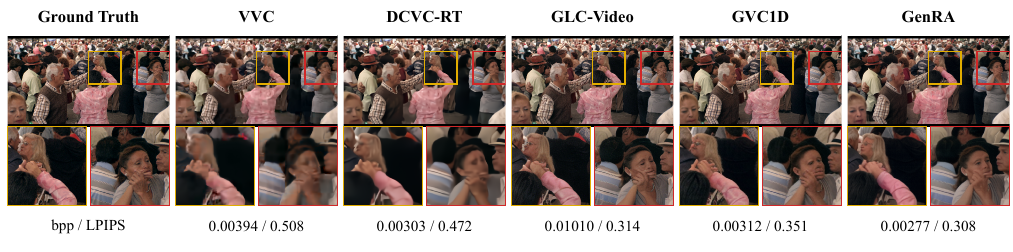}
\caption{\textbf{Qualitative comparison on MCL-JCV.} Labels report bpp / LPIPS. Best viewed when zoomed in.}
\label{fig:qualitative-src14}
\end{figure}
\paragraph{Qualitative comparison.}
\Cref{fig:qualitative-src14} compares reconstructed videos on MCL-JCV~\cite{wang2016mcl}.
At bitrates below $0.005$ bpp, \method recovers facial details and clothing textures.
In contrast, VVC~\cite{bross2021overview} and DCVC-RT~\cite{jia2025towards} produce blurred textures, while GLC-Video~\cite{qi2025glc} and GVC1D~\cite{zheng2026gvc1d} exhibit facial distortions.

\subsection{Effectiveness of Entropy Modeling}
\label{sec:entropy-model-evaluation}

Our entropy model exploits dependencies among video token indices to reduce coding cost.
It combines local and global context for accurate probability estimation and uses grouped prediction for parallel decoding.

\paragraph{Local and Global Context.}
We compare variants without local or global context.
Removing either context increases BD-rate measured using LPIPS and DISTS across all three datasets.
Removing local context increases DISTS-based BD-rate by $13.01\%$--$19.01\%$, while removing global context increases it by $11.02\%$--$17.76\%$.
These results support the complementary roles of local and global context: local context captures spatiotemporal dependencies within windows, while global memory provides cross-window information from previously decoded frames.

\paragraph{Grouped Prediction.}
We compare coding cost and encoding and decoding latency for different group counts with the same frozen tokenizer.
Relative to four groups, two groups reduce decoding latency by approximately $15\%$ but increase bits/token by $0.9\%$--$1.6\%$.
Eight groups reduce bits/token by only $0.7\%$--$0.8\%$ but increase decoding latency by $26\%$--$27\%$.
Although 64 groups achieve the lowest bits/token, decoding takes approximately three times as long as with eight groups.
Thus, increasing the number of groups provides further coding savings at a substantial cost in decoding latency.
We use four groups by default to balance coding efficiency and decoding latency.

\subsection{Joint Rate--Distortion Optimization}
\label{sec:joint-rd-evaluation}

We examine the role of rate gradients in joint training and analyze the effects of rate constraints on token predictability and coding cost.

\begin{figure}[tbp]
\centering
\begin{minipage}[t]{0.485\linewidth}
\centering
\includegraphics[height=4.2cm,width=\linewidth,keepaspectratio]{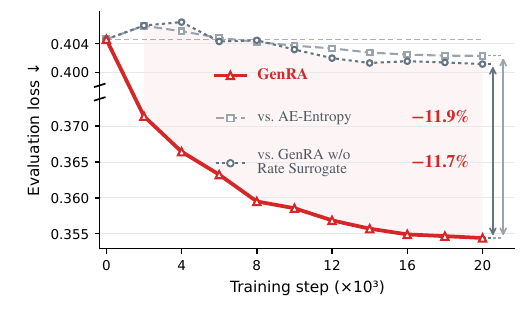}
\caption{Effect of rate gradients on joint rate--distortion training. Percentages show GenRA's evaluation-loss reductions relative to AE-Entropy and GenRA w/o Rate Surrogate at 20K steps.}
\label{fig:stage3-gradient-ablation}
\end{minipage}\hfill
\begin{minipage}[t]{0.485\linewidth}
\centering
\includegraphics[height=4.2cm,width=\linewidth,keepaspectratio]{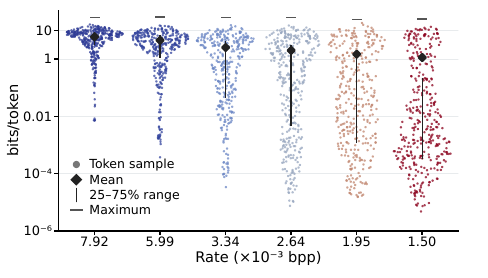}
\caption{\textbf{Coding cost per token on UVG.} The entropy model assigns lower coding costs to tokens under stronger rate constraints.}
\label{fig:token-coding-cost}
\end{minipage}
\end{figure}

\paragraph{Rate surrogate ablation.}
We initialize three variants from the same Stage~II checkpoint.
AE-Entropy freezes the tokenizer and trains only the entropy model.
GenRA w/o Rate Surrogate jointly trains both modules but blocks rate gradients to the encoder.
Full GenRA allows rate gradients to update the encoder.
Figure~\ref{fig:stage3-gradient-ablation} compares these variants on UVG using an evaluation loss of $\mathrm{LPIPS}+20\times\mathrm{bpp}$.
After 20K steps, GenRA achieves evaluation losses $11.9\%$ and $11.7\%$ lower than those of AE-Entropy and GenRA w/o Rate Surrogate, respectively.
This demonstrates the benefit of using coding costs to guide token assignments during joint training.

\paragraph{Token distribution and predictability.}

\begin{figure}[tbp]
\centering
\newsavebox{\temporalcomparisonbox}
\sbox{\temporalcomparisonbox}{\arxivtablestyle
\begin{tabular}{@{}lrrr@{}}
\toprule
\textbf{Method} & \textbf{Rate} & \multicolumn{2}{c}{\textbf{Temporal quality}} \\
\cmidrule(lr){2-2}\cmidrule(lr){3-4}
& bpp $\downarrow$ & $E_{\mathrm{warp}}\,\downarrow$ & CLIP-F $\uparrow$ \\
\midrule
HEVC~\cite{sullivan2012overview} & 0.00636 & 27.29 & 0.9907 \\
VVC~\cite{bross2021overview} & 0.01017 & 25.43 & 0.9921 \\
DCVC-RT~\cite{jia2025towards} & 0.00626 & \textbf{24.74} & 0.9915 \\
\addlinespace[2pt]
GLC-Video~\cite{qi2025glc} & 0.00775 & 128.98 & 0.9884 \\
GVC1D~\cite{zheng2026gvc1d} & 0.00821 & 65.26 & 0.9908 \\
\midrule
\textbf{GenRA (Ours)} & \textbf{0.00568} & 43.24 & \textbf{0.9931} \\
\bottomrule
\end{tabular}
}
\makebox[0.50\linewidth][c]{\raisebox{\dp\temporalcomparisonbox}{\usebox{\temporalcomparisonbox}}}\hfill
\raisebox{-28.105pt}{\includegraphics[width=0.47\linewidth]{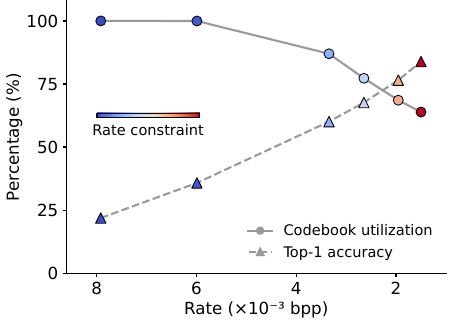}}
\par\vspace{5pt}
\begin{minipage}[t]{0.50\linewidth}
\vspace{0pt}
\captionsetup{type=table,hypcap=false,position=bottom,skip=0pt}
\caption{\textbf{Temporal reconstruction quality on HEVC-B.} Bold values indicate the best result in each column.}
\label{tab:temporal-consistency}
\end{minipage}\hfill
\begin{minipage}[t]{0.47\linewidth}
\vspace{0pt}
\captionsetup{position=bottom,skip=0pt}
\caption{Token predictability under different rate constraints on UVG. Codebook utilization ($\bullet$) and top-1 prediction accuracy ($\blacktriangle$).}
\label{fig:codebook-results}
\end{minipage}
\end{figure}
Figures~\ref{fig:codebook-results} and~\ref{fig:token-coding-cost} show token statistics under different rate constraints.
Stronger rate constraints reduce codebook utilization, which nevertheless remains above $60\%$.
Tokens also become easier to predict, with lower conditional coding costs in Figure~\ref{fig:token-coding-cost} and higher top-1 prediction accuracy in Figure~\ref{fig:codebook-results}.
These results show that joint rate--distortion optimization guides the encoder toward tokens that are more predictable under the entropy model.

\subsection{Temporal Reconstruction Quality}
\label{sec:temporal-evaluation}

Following GNVC-VD~\cite{mao2026generative}, we evaluate temporal consistency using motion-compensated warping error and adjacent-frame CLIP-F.
Table~\ref{tab:temporal-consistency} reports the results on HEVC-B.
\method achieves the highest CLIP-F score of $0.9931$ at the lowest bitrate of $0.00568$ bpp.
Its warping error is $43.24$, compared with $128.98$ for GLC-Video~\cite{qi2025glc} and $65.26$ for GVC1D~\cite{zheng2026gvc1d}.
These results show that \method achieves better temporal consistency than these generative codecs while using fewer bits.

\subsection{Prefix Transmission}
\label{sec:rate-scalability}

\begin{figure}[tbp]
\centering
\includegraphics[width=\linewidth,trim=0 14bp 0 8bp,clip]{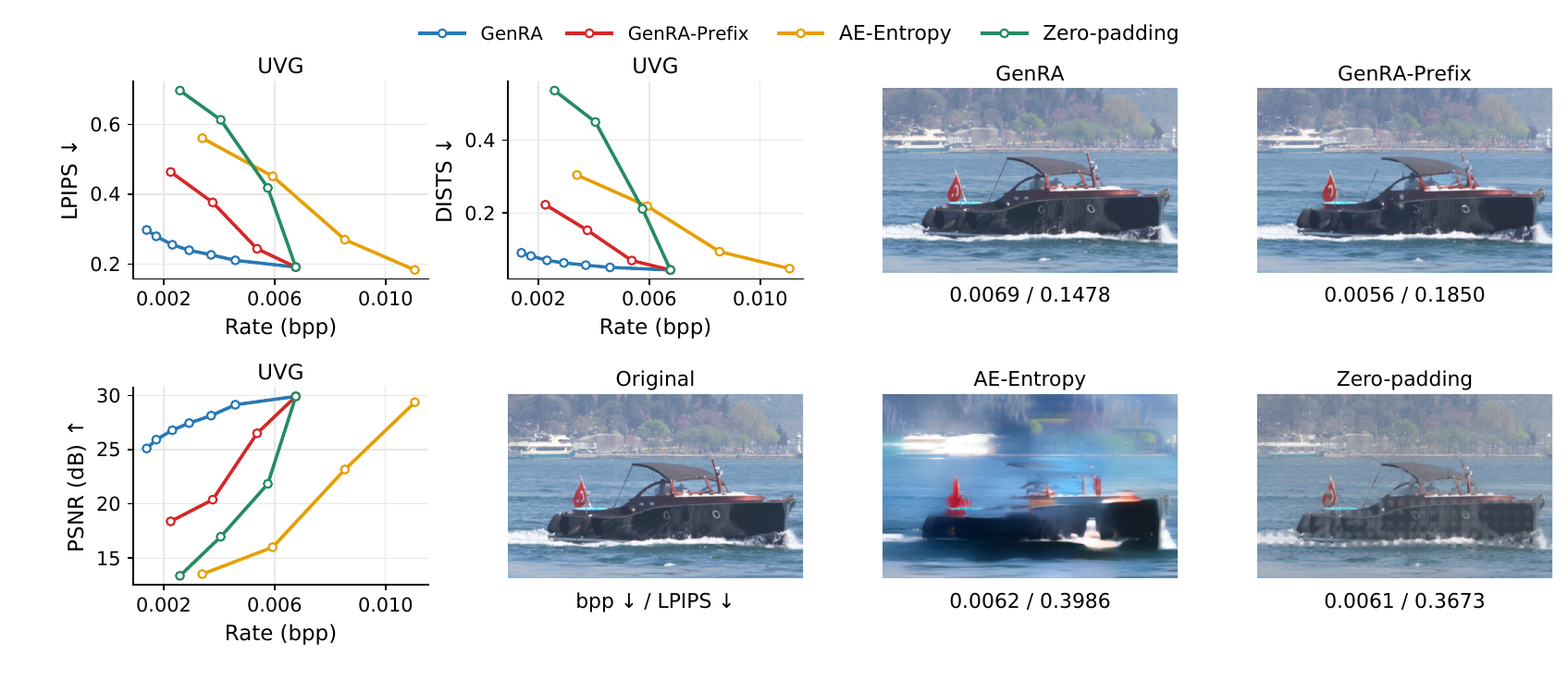}

\caption{\textbf{Test-time rate adjustment via prefix transmission.} LPIPS, DISTS, and PSNR rate--distortion curves on UVG and visual comparisons on Bosphorus. Labels report bpp / LPIPS.}
\label{fig:prefix-results}

\end{figure}

We evaluate test-time rate adjustment by comparing full GenRA with predictive prefix completion, zero padding, and completion using a separately trained tokenizer and entropy model (AE-Entropy).
The prefix and zero-padding variants use the same $\lambda=7.2$ checkpoint and transmit the first $k\in\{1,2,3\}$ spatial groups.
Figure~\ref{fig:prefix-results} shows that GenRA-Prefix achieves lower LPIPS and DISTS while using fewer bits than both controls at each partial-transmission level.
Reconstruction quality decreases as fewer groups are transmitted.
These results support the use of the entropy model as a completion prior for test-time bitrate adjustment without retraining.

\subsection{Computational Complexity}
\label{sec:computational-efficiency}

Table~\ref{tab:entropy-latency} compares end-to-end encoding and decoding latency on UVG at three resolutions: $1920\times1080$, $1280\times720$, and $854\times480$.
At 1080p, \method requires $252.91$ ms/frame for encoding and $159.03$ ms/frame for decoding.
Compared with GVC1D~\cite{zheng2026gvc1d}, \method is slightly slower to encode at 1080p and 720p but faster to decode at all three resolutions.
GLC-Video~\cite{qi2025glc} and DCVC-FM~\cite{li2024neural} remain faster, while \method achieves much better perceptual reconstructions.

\begin{table}[tbp]
\arxivtablestyle
\centering
\sbox{\arxivtablebox}{%
\begin{tabular}{@{}lrrrrrrr@{}}
\toprule
\textbf{Method} & \textbf{Params} & \multicolumn{2}{c}{\textbf{1080p}} & \multicolumn{2}{c}{\textbf{720p}} & \multicolumn{2}{c}{\textbf{480p}} \\
\cmidrule(lr){3-4}\cmidrule(lr){5-6}\cmidrule(lr){7-8}
& (B) & Enc. $\downarrow$ & Dec. $\downarrow$ & Enc. $\downarrow$ & Dec. $\downarrow$ & Enc. $\downarrow$ & Dec. $\downarrow$ \\
\midrule
DCVC-FM~\cite{li2024neural} & 0.045 & \textbf{118.57} & \textbf{97.14} & \textbf{55.56} & \textbf{46.10} & \textbf{29.88} & \textbf{24.85} \\
GLC-Video~\cite{qi2025glc} & 0.226 & 208.36 & 130.18 & 102.13 & 63.74 & 50.67 & 31.71 \\
GVC1D~\cite{zheng2026gvc1d} & 1.772 & 242.74 & 177.28 & 109.83 & 82.13 & 81.10 & 61.32 \\
\midrule
\textbf{GenRA (Ours)} & 2.629 & 252.91 & 159.03 & 117.63 & 73.17 & 62.44 & 40.08 \\
\bottomrule
\end{tabular}%
}
\begin{minipage}{\wd\arxivtablebox}
\caption{\textbf{Model size and end-to-end coding latency on UVG.} Enc./Dec. are measured in ms/frame on an NVIDIA H100 (FP16).}
\label{tab:entropy-latency}
\noindent\usebox{\arxivtablebox}
\end{minipage}

\end{table}

\FloatBarrier
\section{Conclusion}
We presented \method, a framework that adapts pretrained video generative representations for direct compression without an additional learned codec.
Experiments on UVG~\cite{mercat2020uvg}, MCL-JCV~\cite{wang2016mcl}, and HEVC-B show superior perceptual rate--distortion performance, demonstrating the effectiveness of representation adaptation for video compression.
However, the large video VAE remains the main latency bottleneck.
Future work could distill \method into a smaller model to reduce coding latency while retaining its rate--distortion performance.

\setlength{\bibsep}{2pt plus 0.5pt}
\bibliographystyle{ieeenat_fullname}
\bibliography{references}
\end{document}